# PsychoLex: Unveiling the Psychological Mind of Large Language Models

Mohammad Amin Abbasi[1], Farnaz Sadat Mirnezami[2], Hassan Naderi[1]

[1]Department of Computer Engineering, Iran University of Science and Technology, Tehran, Iran
[2]Department of Computer Engineering, University of Guilan, Rasht, Iran

m_abbasi1378@comp.iust.ac.ir
farnaz.mirnezami@gmail.com
naderi@iust.ac.ir

## Abstract

This paper explores the intersection of psychology and artificial intelligence through the development and evaluation of specialized Large Language Models (LLMs). We introduce PsychoLex [1], a suite of resources designed to enhance LLMs' proficiency in psychological tasks in both Persian and English. Key contributions include the PsychoLexQA dataset for instructional content and the PsychoLexEval dataset for rigorous evaluation of LLMs in complex psychological scenarios. Additionally, we present the PsychoLexLLaMA model, optimized specifically for psychological applications, demonstrating superior performance compared to general-purpose models. The findings underscore the potential of tailored LLMs for advancing psychological research and applications, while also highlighting areas for further refinement. This research offers a foundational step towards integrating LLMs into specialized psychological domains, with implications for future advancements in AI-driven psychological practice.

## 1 Introduction

The rise of Large Language Models (LLMs) has significantly advanced artificial intelligence (AI), providing remarkable capabilities in natural language processing and understanding (Guo et al., 2023; Minaee et al., 2024; Wu et al., 2023). These models have shown proficiency in generating human-like text, translating languages, and engaging in sophisticated dialogues (Agrawal, 2023). However, as users increasingly rely on LLMs for psychological and therapeutic questions (Lai et al., 2023), the limitations of these models in specialized domains have become apparent. Notably, there is a critical absence of datasets designed to evaluate and enhance LLMs' performance in the field of psychology.

Despite considerable progress in general AI research, the integration of psychological expertise into LLMs remains underdeveloped. Existing methodologies often lack the depth required to understand and respond accurately to complex psychological inquiries. Moreover, the field is hindered by the lack of comprehensive datasets that include not only questions and answers but also instructional content tailored to psychological contexts. This gap is significant because it restricts the practical applications of LLMs in psychological research, therapy, and education, where nuanced and precise information is essential.

Our research seeks to address this gap by introducing PsychoLex, a suite of resources and models specifically designed for psychological applications in both Persian and English. The primary objectives of this study are to develop and evaluate specialized datasets, namely PsychoLexQA and PsychoLexEval, and to introduce PsychoLexLLaMA, an LLM developed for psychologyical tasks. These contributions include: (i) PsychoLexQA, which provides comprehensive instructional content and detailed questions and answers to enhance LLM training; (ii) PsychoLexEval, a multiple-choice question and answer (MCQA) dataset designed for rigorous evaluation of LLMs in psychological contexts, ensuring they can handle complex psychological queries accurately and contextually;(iii) PsychoLexLLaMA, which improves the performance of LLMs in psychological tasks through continual pre-training and fine-tuning of LLaMA 3.1.(Dubey et al., 2024) Together, these contributions aim to provide robust solutions to existing challenges, enhancing the accuracy and relevance of AI-driven psychological tools and paving the way for future advancements in integrating AI with psychological practice.

---

[1] https://huggingface.co/collections/aminabbasi/psycholex-66b64e3768da519596e49de9

The structure of this paper is organized as follows: Section 2 reviews related work in LLMs and their applications in psychology. Section 3 details the datasets developed for this study, including their creation and validation processes. Section 4 discusses the development and fine-tuning of the PsychoLexLLaMA model. Section 5 presents the evaluation methodology and results, comparing PsychoLexLLaMA with other state-of-the-art models. Section 6 provides a comprehensive discussion of the findings, and Section 7 concludes the paper with insights into future research directions and potential applications.

By exploring the intersection of AI and psychology, this paper aims to unveil the psychological capabilities of LLMs and demonstrate their potential to advance both fields significantly.

## 2 Related Works

In this section, we review existing research that benchmarks the capabilities of large language models (LLMs) in Persian, followed by studies that explore the integration of LLMs into psychological research and applications. This dual-focus review establishes the context for our work, emphasizing both the linguistic challenges specific to Persian and the broader implications of applying LLMs in the field of psychology.

### 2.1 Benchmarking Large Language Models for Persian

Recent advancements in large language models (LLMs), particularly ChatGPT, have generated significant interest in their evaluation across various languages and tasks. ChatGPT's performance on various Persian natural language processing tasks is evaluated by Abaskohi et al. (2024). they present a comprehensive evaluation of large language models (LLMs) for the Persian language, focusing on models like GPT-3.5-turbo (OpenAI, 2023a), GPT-4 (OpenAI, 2023b), and OpenChat-3.5. This study, which is the first extensive benchmarking effort for Persian, aims to address the challenges posed by Persian as a low-resource language with unique linguistic features. The evaluation covers a broad range of natural language processing (NLP) tasks, including sentiment analysis, question answering, natural language inference, and translation. the study highlights the model's superior performance in multiple-choice questions(MCQs) related to math and general knowledge from the ParsiNLU dataset (Khashabi et al., 2020). These benchmarks are particularly important for assessing the models' reasoning capabilities in Persian. While ChatGPT-4 excels across several benchmarks, its application in psychology has not been tested, underscoring a critical area for future research.

### 2.2 Khayyam Challenge (PersianMMLU)

Recent advancements have focused on optimizing the performance of Large Language Models (LLMs). The PersianMMLU (Ghahroodi et al., 2024) is particularly significant as it concentrates on the Persian language capabilities of these models. It evaluates their proficiency in answering multiple-choice questions across diverse fields such as mathematics, science, logic, and intelligence testing. This comprehensive evaluation involved advanced models like GPT-3.5, GPT-4(OpenAI, 2023b), Aya (Ustun et al., 2024), PersianMind(Rostami et al., 2024), mT0 (Muennighoff et al., 2023), mGPT (Shliazhko et al., 2022), and Claude3-haiku (Anthropic, 2024). The study utilized a robust dataset derived from Iran's national university entrance exams and educational assessments. While GPT-4 emerged as the superior model, its efficacy in psychological applications remains untested. This gap highlights the necessity of our current research, which aims to specifically evaluate the performance of LLMs in psychology-related scenarios.

### 2.3 Using large language models in psychology

Dubey et al. (2024) explores the integration of LLMs, particularly GPT-3 and GPT-4, into psychological research practices. These models' adeptness at text generation, dialogue engagement, persona simulation, and information synthesis provides innovative approaches to studying various psychological subfields. The primary aim is to evaluate the extent to which LLMs can enrich psychological research methodologies. Despite their potential, LLMs often fall short in delivering contextually accurate advice consistently. This study highlights the importance of refining LLMs through fine-tuning and reinforcement learning from human feedback to ensure their practical efficacy in real-world psychological settings. The extensive datasets used to train these models, encompassing diverse sources of human language data, are aimed at tailoring LLMs to better serve both theoretical and applied psychology.

### 2.4 Exploring the Frontiers of LLMs in Psychological Applications

The application of Artificial Intelligence (AI), especially large language models (LLMs), is revolutionizing psychological research. A study by Ke et al. (2024) underscores significant advances in language models and their profound impact on the field of psychology. LLMs like OpenAI's ChatGPT facilitate various research activities, including literature reviews, hypothesis formulation, experiment design, data analysis, and scholarly writing across several psychological domains such as cognitive, behavioral, clinical, educational, developmental, and social psychology. While these models offer substantial benefits, the review also delineates key technical and ethical challenges, including data privacy concerns and inherent limitations of LLMs. The authors advocate for the careful integration of these technologies in psychological research to enhance our understanding of the human mind and improve the methodologies employed in psychological studies.

## 3 Dataset

This section outlines the datasets developed to investigate the application of large language models (LLMs) in psychology. We detail the creation and utilization of three pivotal datasets: the foundational pretraining data, the PsychoLexQA dataset for instructional content, and the PsychoLexEval dataset for evaluating model comprehension and performance.

### 3.1 Pretraining Data

For the pretraining phase, we employed "Introduction to Psychology" by Hilgard (1953), a seminal textbook noted for its comprehensive insights into psychology. This text was used in both its Persian and English versions to establish a bilingual foundation for our models. The dataset comprised approximately 1.3 million tokens, offering a rich and diverse corpus that spans a broad spectrum of psychological topics. This extensive pretraining data enabled our models to develop a deep understanding of essential psychological concepts and terminology, facilitating their application in both Persian and English contexts.

### 3.2 PsychoLexQA

For the instructional dataset, we adopted two distinct methodologies to generate detailed and comprehensive instructional content in both Persian and English. Appendix A demonstrates two examples of the PsychoLexQA dataset.

#### 3.2.1 Document-Based Instructions

The first method involved extracting instructional content from "Introduction to Psychology" in both languages. This process was automated using the GPT-4o model, where paragraphs from the textbook were analyzed to grasp key concepts. For each paragraph, the model generated a series of questions and answers aimed at testing material comprehension. Each question was crafted to be clear and precise, with detailed answers provided to ensure a thorough understanding of the discussed psychological concepts. Paragraphs lacking sufficient content for question generation were identified and noted. This method resulted in a dataset containing 7,055 entries.

#### 3.2.2 Self-Instruct

The second method focused on creating structured instructional tasks for various psychological subcategories in both Persian and English. This involved defining tasks such as "Case Study Analysis", "Experiment Design", and "Data Interpretation" across different psychological subfields like Clinical Psychology and Cognitive Psychology. For each task and subcategory combination, detailed instructions were generated, including task descriptions, optional inputs, and expected outputs. These tasks were presented in a bilingual format, accommodating both Persian and English speakers. The dataset created using GPT-4o comprises a total of 3,001 rows, ensuring extensive coverage of psychological topics.

## 3.3 PsychoLexEval

The PsychoLexEval dataset, a multiple-choice question and answer (MCQA) format in both Persian and English, is designed to assess the comprehension and performance of LLMs in psychology. This section will describe the data collection and review process, the methods employed to ensure quality and compliance, and the broad scope and coverage of this MCQA dataset.

### 3.3.1 Data Collection

To construct this dataset, we compiled questions from multiple significant sources: (1) Graduate Entrance Exams: questions from psychology entrance exams (2014-2024) that cover advanced topics; (2) Employment Exams: questions from various job tests, including both specialized and general psychology; (3) Online Sources: Questions from trusted psychology test websites; (4) GPT-4 Generated Content: questions from Psychology books, covering a wide range of topics.

### 3.3.2 Filtering and Review

To ensure high quality and legal compliance, we implemented rigorous filtering and review processes for the dataset. Initially, a human review was conducted where a sample of questions was meticulously scrutinized by experts. This step was crucial to ensure that each question was relevant, complete, and clearly articulated. During this phase, we specifically retained only those questions that had exactly four answer options, ensuring consistency and clarity in the evaluation process. Additionally, to avoid any legal complications, we carefully removed any content that potentially violated copyright laws. This step was essential to maintain the integrity of the dataset and ensure that all included materials were legally compliant for use in our research and broader academic dissemination. These measures collectively reinforced the dataset's reliability and adherence to legal standards, providing a robust foundation for evaluating large language models within psychological contexts.

### 3.3.3 Scope and Coverage

The PsychoLexEval dataset is meticulously designed to evaluate the comprehension and performance of large language models (LLMs) in the field of psychology, ensuring comprehensive coverage across a wide spectrum of psychological fields. It includes general psychology, focusing on basic concepts; developmental psychology, which examines human growth and cognitive development; and clinical psychology, addressing the diagnosis and treatment of mental disorders. Additionally, the dataset encompasses psychometrics, highlighting methods of measuring psychological attributes; cognitive tests for assessing intelligence and aptitude; and industrial and organizational psychology, which looks at behavior and productivity in the workplace.

Further expanding its breadth, the dataset covers social psychology, which explores social behaviors and group dynamics; educational psychology, focusing on learning processes and teaching methods; and the needs of exceptional children with special requirements. It also integrates key concepts from "Introduction to Psychology," covering a range of fundamental topics including biological foundations, sensory processes, learning, memory, motivation, emotion, intelligence, personality, and psychological disorders.

By providing such diverse and extensive content, the PsychoLexEval dataset serves as an invaluable resource for researchers in psychology and artificial intelligence. It equips them with a robust tool to deepen insights into psychological phenomena and advance the field by effectively evaluating the capabilities of LLMs across various psychology domains. The dataset comprises 3,430 rows. Appendix A shows an example of the PsychoLexEval dataset.

## 4 PsychoLexLLaMA

In this section, we detail the development of PsychoLexLLaMA, a specialized large language model (LLM) designed explicitly for psychology. Our goal was to surpass the performance of general-purpose models by optimizing our model to require minimal data and hardware resources. We utilized the Transformers[2] library for model development The process of constructing our model is illustrated in Appendix A.

### 4.1 Continuous Pre-Training

For continuous pre-training (Zhou et al., 2024), we employed the LoRA technique (Hu et al., 2021) on the bilingual texts of "Introduction to

---
[2] https://github.com/huggingface/transformers

Psychology" by Hilgard. This foundational work was processed in both Persian and English, leveraging the established pretraining data. We utilized LLaMA 3.1(Dubey et al., 2024) as our base models in two configurations: 8B and 70B. This stage was critical for aligning the base models with psychological content, thereby enhancing their understanding and application of complex psychological concepts efficiently. The pre-training for the 8B model took 8 minutes using a single A100 80GB GPU, while the 70B model required 41 minutes on two A100 80GB GPUs. Table 1 provides a detailed overview of the LoRA training configurations used during this phase.

| Lr | Rank | Alpha | Dropout |
|---|---|---|---|
| 1e-5 | 8 | 16 | 0.0 |

Table 1: LoRA training configurations

### 4.2 Supervised Fine-Tuning

The supervised fine-tuning phase was essential for tailoring our models to meet the specific demands of psychological analysis. Utilizing the PsychoLexQA dataset, which includes both instructional content and a comprehensive set of questions and answers, we applied the LoRA technique to further train the pre-trained models. This phase was pivotal in refining the models' abilities to interpret and respond accurately to intricate psychological queries and scenarios within the dataset. The supervised fine-tuning for the 8B model took 22 minutes using a single A100 GPU, while the 70B model required 32 minutes on two A100 GPUs. The LoRA training configurations used during this phase were the same as those in the continuous pre-training.

### 4.3 Linear Weight Combination

To bolster the final model's robustness and preserve the integrity of previous training advances, we implemented a linear weight combination strategy. This involved merging the weights of the LLaMA 3.1 Instruct model with our continuously pre-trained and finely-tuned models. Each model contributed 50% of its weight to the final composite. This method synergistically combined the foundational capabilities of LLaMA with our newly developed psychological expertise, producing a balanced and potent tool adept at handling sophisticated psychological inquiries.

Through these meticulous steps, PsychoLexLLaMA has been meticulously tailored to meet the unique needs of psychological applications. It stands as a robust resource for researchers and practitioners in both psychology and artificial intelligence, providing a reliable platform for further explorations and advancements in these fields. The next sections will evaluate PsychoLexLLaMA's performance in detail, comparing it with other models to underscore its enhanced capabilities in the realm of psychological research and practice.

## 5 Evaluation

In this study, we conducted a comprehensive evaluation of various language models that operate in both Persian and English, focusing on their ability to understand and accurately respond to psychological questions. The models assessed include include Qwen2 (Yang et al., 2024), Aya-23 (Aryabumi et al., 2024), Phi-3 (Abdin et al., 2024), Llama-3, Llama-3.1(Dubey et al., 2024), Gemma 1.1 (Team et al., 2024), command-r, PersianLLaMA (Abbasi et al., 2023), PersianMind (Rostami et al., 2024b), and PsychoLexLLaMA. Our focus on open-source models was intended to enhance the accessibility and reproducibility of our findings. The generation configuration for all the LLMs evaluated is consistent across the experiments and is detailed in Table 2.

| Temp | Max new tokens | top p | Do sample |
|---|---|---|---|
| 0.01 | 16 | 0.9 | True |

Table 2: Generation configurations for all evaluated LLMs.

### 5.1 Zero-shot Setting

In the zero-shot setting, models were tested without any prior contextual examples, relying solely on their pre-existing knowledge. This setting evaluated the models' intrinsic ability to generate accurate responses based solely on their training.

### 5.2 One-shot Setting

The one-shot setting involved presenting each model with a single relevant example before it answered a question. This setting was used to assess the impact of a minimal context on the accuracy of the models, providing insights into their ability to leverage new information quickly

### 5.3 Five-shot Setting

In the five-shot setting, models were given five related examples before responding to questions. This scenario tested the models' capacity to utilize more extensive contextual information to enhance their accuracy, offering a deeper understanding of their learning capabilities.

### 5.4 Evaluation Metric

The effectiveness of each model across the zero-shot, one-shot, and five-shot settings was measured using accuracy as the primary metric. Accuracy was defined as the proportion of correct answers provided by the models relative to the total number of questions posed. This rigorous evaluation approach allowed us to discern the strengths and weaknesses of each model in processing and understanding psychological content comprehensively.

Through these methodical evaluations, we aimed to illustrate the varying capabilities of each model under different contextual conditions. This analysis not only sheds light on how models adapt to incremental information but also highlights their potential applicability in psychological settings, where understanding nuanced human behavior is crucial.

## 6 Results

This section outlines the outcomes of our evaluation of selected large language models (LLMs) using the PsychoLexEval dataset in both Persian and English. The primary focus was on assessing the models' proficiency in understanding and responding to psychological questions.

Tables 3 and 4 illustrate the accuracy results of the models on the PsychoLexEval dataset for Persian and English, respectively. These tables quantify how effectively each model comprehends and addresses psychology-related questions across languages.

### 6.1 Discussion

The results from Tables 3 and 4 provide significant insights into the performance of various LLMs, showcasing their competencies in both Persian English. Notably, these findings highlight the influence of model architecture and parameter size on handling specialized tasks, such as interpreting and responding to psychology-related questions.

#### 6.1.1 Performance Trends Across Models

The data reveal substantial variability in performance across models and settings. For instance, the Llama-3.1 Instruct with 70B parameters exhibits superior performance in all scenarios, suggesting a positive correlation between larger parameter sizes and enhanced comprehension and response accuracy. This trend is consistent in the English data, where models with larger parameters, such as Llama-3.1 Instruct 70B, also demonstrate robust performance, especially in zero-shot and five-shot settings.

Conversely, models with fewer parameters sometimes perform well in lower-shot settings but typically exhibit decreased performance as the complexity of tasks increases. For example, the Qwen2 Instruct with 7B parameters faces greater challenges in the Persian context than in English, potentially indicating linguistic or dataset-specific hurdles that are more effectively managed by larger models.

#### 6.1.2 Language-Specific Observations

Our evaluation underscores distinct language-specific differences. In Persian, the increase in model accuracy from zero to five shots is more marked, indicating that Persian language models significantly benefit from added context. Conversely, English language models tend to have higher baseline performances, likely reflecting the advantages of more extensive pre-training datasets available in English.

#### 6.1.3 Impact of Training and Fine-Tuning

The results particularly underscore the critical importance of targeted training and fine-tuning, as seen with the PsychoLexLLaMA models. Designed to surpass its predecessor, Llama 3.1, the 70B PsychoLexLLaMA occasionally does not reach its ambitious targets but consistently matches or exceeds the performance of the original Llama 3.1 model. This consistency indicates that while specific enhancements did not universally lead to improvements, they significantly bolstered the model's capabilities. The 70B version, with its vast parameter count, possesses the capacity to acquire a broader knowledge base, making it challenging to add new knowledge without forgetting previously learned information. Consequently, fine-tuning such a large model demands considerably more data to achieve better outcomes due to its complexity.

| Model | # Param | Accuracy | | | |
|---|---|---|---|---|---|
| | | 0-shot | 1-shot | 5-shot | Avg |
| Qwen2 Instruct | 7B | 03.55 | 06.18 | 08.63 | 6.12 |
| Gemma 1.1 it | 7B | 43.07 | 40.68 | 27.57 | 37.11 |
| PersianMind | 7B | 35.78 | 35.96 | 24.63 | 32.12 |
| Aya-23 | 8B | 39.64 | 41.42 | 27.02 | 36.03 |
| Llama-3 Instruct | 8B | 33.88 | 10.66 | 34.49 | 26.34 |
| Llama-3.1 Instruct | 8B | 45.89 | 41.36 | 35.78 | 41.01 |
| PsychoLexLLaMA-pretrain-sft | 8B | 47.30 | 43.13 | 46.61 | 45.68 |
| **PsychoLexLLaMA-average** | **8B** | **48.52** | **41.97** | **47.05** | **45.85** |
| PersianLLaMA | 13B | 20.13 | 18.52 | 19.89 | 19.51 |
| Aya-23 | 35B | 21.07 | 10.47 | 22.69 | 18.08 |
| c4ai-command-r-v01 | 35B | 35.96 | 21.75 | 46.20 | 34.64 |
| Llama-3 Instruct | 70B | 19.54 | 09.31 | 0.5 | 9.78 |
| **Llama-3.1 Instruct** | **70B** | **70.34** | **67.83** | **70.40** | **69.52** |
| PsychoLexLLaMA-pretrain-sft | 70B | 67.79 | 45.34 | 68.07 | 60.4 |
| PsychoLexLLaMA-average | 70B | 65.84 | 53.06 | 69.66 | 62.85 |
| Qwen2 Instruct | 72B | 31.37 | 05.82 | 50.3 | 29.16 |

Table 3 : Accuracy of LLMs on the PsychoLexEval dataset in Persian.

| Model | # Param | Accuracy | | | |
|---|---|---|---|---|---|
| | | 0-shot | 1-shot | 5-shot | Avg |
| Qwen2 Instruct | 7B | 89.31 | 42.74 | 83.76 | 71.94 |
| Gemma 1.1 it | 7B | 84.75 | 55.06 | 65.86 | 68.56 |
| Aya-23 | 8B | 73.62 | 33.80 | 77.05 | 61.49 |
| Llama-3 Instruct | 8B | 85.77 | 78.57 | 68.22 | 77.52 |
| Llama-3.1 Instruct | 8B | 88.97 | 89.25 | 87 | 88.41 |
| PsychoLexLLaMA-pretrain-sft | 8B | 88.97 | 81.21 | 62.03 | 77.4 |
| **PsychoLexLLaMA-average** | **8B** | **90.10** | **89.03** | **90.04** | **89.72** |
| Aya-23 | 35B | 81.32 | 79.02 | 82 | 80.78 |
| c4ai-command-r-v01 | 35B | 87 | 78.06 | 75.08 | 80.05 |
| Llama-3 Instruct | 70B | 90.55 | 88.58 | 76.77 | 85.3 |
| **Llama-3.1 Instruct** | **70B** | **93.02** | **92.63** | **92.1** | **92.58** |
| PsychoLexLLaMA-pretrain-sft | 70B | 91.45 | 90.24 | 90.85 | 90.85 |
| PsychoLexLLaMA-average | 70B | 92.13 | 91.85 | 91.87 | 91.95 |
| Qwen2 Instruct | 72B | 91.11 | 73.79 | 92.29 | 85.73 |

Table 4 : Accuracy of LLMs on the PsychoLexEval dataset in English.

In contrast, the 8B version of PsychoLexLLaMA often outperforms larger models, suggesting that precise, domain-specific fine-tuning can yield remarkable effectiveness, even with fewer parameters. This success highlights the potential of smaller models, particularly when equipped with tailored enhancements for specific applications like psychological evaluations.

The varying impacts of scaling between the 8B and 70B versions suggest that while larger models possess a broad knowledge base enhancing their general performance, strategic fine-tuning is crucial for maximizing efficacy in specialized domains. This observation encourages further research into training strategies that optimize both large and small models for specific tasks, ensuring that they not only retain previous knowledge but also effectively integrate new information.

## 7 Conclusion

This study has significantly advanced our understanding of how large language models (LLMs) can be effectively tailored for applications within psychology. Through the integration of specialized psychological content, the development of the PsychoLexQA and PsychoLexEval datasets, and the creation of the PsychoLexLLaMA model, we have demonstrated the substantial benefits of targeted model training and fine-tuning.

Our findings indicate that specific pretraining and fine-tuning strategies substantially enhance the performance of LLMs in psychological settings, underscoring the critical role of thoughtful model architecture and training approaches. Notably, while larger models typically show strong performance, our results reveal that even smaller models can achieve exceptional outcomes when subjected to precise, domain-specific adjustments. This suggests a scalable potential for LLMs in psychological applications that can be adapted to different contexts and constraints.

In conclusion, this research not only sheds light on the current capabilities and challenges of using LLMs in psychology but also sets a foundation for future work. It encourages ongoing refinement of these models to improve their relevance and accuracy, thereby enhancing their utility in real-world psychological applications. Moving forward, we anticipate that continued advancements in model training methodologies and evaluation strategies will drive significant progress in the field, making LLMs an indispensable tool in the arsenal of psychological research and practice.

## Limitations

This research has demonstrated the potential of customizing large language models (LLMs) for psychological applications. However, it is crucial to recognize several limitations that could impact the scope and applicability of our findings. The PsychoLexQA and PsychoLexEval datasets, fundamental to our study, inherently contain biases due to the selection of textual materials and question design. These biases may restrict the generalizability of our results to broader psychological contexts and populations. Additionally, the reliance on freely licensed sources limits the diversity and depth of psychological topics explored, omitting valuable content protected by copyright laws.

Another significant constraint is the dependence on sophisticated hardware for model training. The need for high-performance GPUs poses a considerable barrier, particularly for researchers with limited access to such resources, affecting both the replicability of our results and the broader research community's ability to engage with and expand upon our work. Moreover, while this study aims to enhance model performance with minimal data and hardware resources, achieving optimal efficiency under these constraints remains a challenge. Balancing resource conservation with model capability often requires compromises that may detract from the models' utility in practical psychological applications.

By addressing these limitations, future research can focus on broadening the diversity of training data and developing more resource-efficient modeling techniques, thereby enhancing the practical deployment of LLMs in psychology and related fields.

**References**

Abaskohi, A., Baruni, S., Masoudi, M., Abbasi, N., Babalou, M. H., Edalat, A., Kamahi, S., Sani, S. M., Naghavian, N., Namazifard, D., Sadeghi, P., & Yaghoobzadeh, Y. (2024). Benchmarking Large Language Models for Persian: A Preliminary Study Focusing on ChatGPT. *ArXiv, abs/2404.02403*.

Abbasi, M. A., Ghafouri, A., Firouzmandi, M., Naderi, H., & Minaei-Bidgoli, B. (2023). PersianLLaMA: Towards Building First Persian Large Language Model. *ArXiv, abs/2312.15713*.

Abdin, M., Jacobs, S. A., Awan, A. A., Aneja, J., Awadallah, A., Awadalla, H., Bach, N., Bahree, A., Bakhtiari, A., & Behl, H. (2024). Phi-3 technical report: A highly capable language model locally on your phone. *arXiv preprint arXiv:2404.14219*.

Agrawal, S. (2023). Are LLMs the Master of All Trades? : Exploring Domain-Agnostic

Reasoning Skills of LLMs. *ArXiv*, *abs/2303.12810*.

Anthropic. (2024,

). The Claude 3 Model Family: Opus, Sonnet, Haiku.

Aryabumi, V., Dang, J., Talupuru, D., Dash, S., Cairuz, D., Lin, H., Venkitesh, B., Smith, M., Marchisio, K., & Ruder, S. (2024). Aya 23: Open weight releases to further multilingual progress. *arXiv preprint arXiv:2405.15032*.

Dubey, A., Jauhri, A., Pandey, A., Kadian, A., Al-Dahle, A., Letman, A., Mathur, A., Schelten, A., Yang, A., Fan, A., Goyal, A., Hartshorn, A., Yang, A., Mitra, A., Sravankumar, A., Korenev, A., Hinsvark, A., Rao, A., Zhang, A., . . . Zhao, Z. (2024). The Llama 3 Herd of Models.

Ghahroodi, O., Nouri, M., Sanian, M. V., Sahebi, A., Dastgheib, D., Asgari, E., Baghshah, M. S., & Rohban, M. H. (2024). Khayyam Challenge (PersianMMLU): Is Your LLM Truly Wise to The Persian Language? *ArXiv*, *abs/2404.06644*.

Guo, Z., Jin, R., Liu, C., Huang, Y., Shi, D., Supryadi, Yu, L., Liu, Y., Li, J., Xiong, B., & Xiong, D. (2023). Evaluating Large Language Models: A Comprehensive Survey. *ArXiv*, *abs/2310.19736*.

Hu, J. E., Shen, Y., Wallis, P., Allen-Zhu, Z., Li, Y., Wang, S., & Chen, W. (2021). LoRA: Low-Rank Adaptation of Large Language Models. *ArXiv*, *abs/2106.09685*.

Ke, L., Tong, S., Cheng, P., & Peng, K. (2024). Exploring the Frontiers of LLMs in Psychological Applications: A Comprehensive Review. *ArXiv*, *abs/2401.01519*.

Khashabi, D., Cohan, A., Shakeri, S., Hosseini, P., Pezeshkpour, P., Alikhani, M., Aminnaseri, M., Bitaab, M., Brahman, F., Ghazarian, S., Gheini, M., Kabiri, A., Mahabadi, R. K., Memarrast, O., Mosallanezhad, A., Noury, E., Raji, S., Rasooli, M. S., Sadeghi, S., . . . Yaghoobzadeh, Y. (2020). ParsiNLU: A Suite of Language Understanding Challenges for Persian. *Transactions of the Association for Computational Linguistics*, *9*, 1147-1162.

Lai, T., Shi, Y., Du, Z., Wu, J., Fu, K., Dou, Y., & Wang, Z. (2023). Psy-LLM: Scaling up Global Mental Health Psychological Services with AI-based Large Language Models. *ArXiv*, *abs/2307.11991*.

Minaee, S., Mikolov, T., Nikzad, N., Chenaghlu, M. A., Socher, R., Amatriain, X., & Gao, J. (2024). Large Language Models: A Survey. *ArXiv*, *abs/2402.06196*.

Muennighoff, N., Wang, T., Sutawika, L., Roberts, A., Biderman, S., Scao, T. L., Bari, M. S., Shen, S., Yong, Z.-X., Schoelkopf, H., Tang, X., Radev, D. R., Aji, A. F., Almubarak, K., Albanie, S., Alyafeai, Z., Webson, A., Raff, E., & Raffel, C. (2023). Crosslingual Generalization through Multitask Finetuning. Annual Meeting of the Association for Computational Linguistics,

OpenAI. (2023a). *Gpt-3.5*. https://www.openai.com/. Accessed: 2023-06-13.

OpenAI. (2023b). *Gpt-4 technical report*. https://arxiv.org/abs/2303.08774

Rostami, P., Salemi, A., & Dousti, M. J. (2024). PersianMind: A Cross-Lingual Persian-English Large Language Model. *ArXiv*, *abs/2401.06466*.

Shliazhko, O., Fenogenova, A., Tikhonova, M., Mikhailov, V., Kozlova, A., & Shavrina, T. (2022). mGPT: Few-Shot Learners Go Multilingual. *Transactions of the Association for Computational Linguistics*, *12*, 58-79.

Team, G., Mesnard, T., Hardin, C., Dadashi, R., Bhupatiraju, S., Pathak, S., Sifre, L., Rivière, M., Kale, M. S., & Love, J. (2024). Gemma: Open models based on gemini research and technology. *arXiv preprint arXiv:2403.08295*.

Ustun, A., Aryabumi, V., Yong, Z.-X., Ko, W.-Y., D'souza, D., Onilude, G., Bhandari, N., Singh, S., Ooi, H.-L., Kayid, A., Vargus, F., Blunsom, P., Longpre, S., Muennighoff, N., Fadaee, M., Kreutzer, J., & Hooker, S. (2024). Aya Model: An Instruction Finetuned Open-Access Multilingual Language Model. *ArXiv*, *abs/2402.07827*.

Wu, Q., Bansal, G., Zhang, J., Wu, Y., Li, B., Zhu, E., Jiang, L., Zhang, X., Zhang, S., Liu, J., Awadallah, A. H., White, R. W., Burger, D., & Wang, C. (2023). AutoGen: Enabling Next-Gen LLM Applications via Multi-Agent Conversation.

Yang, A., Yang, B., Hui, B., Zheng, B., Yu, B., Zhou, C., Li, C., Li, C., Liu, D., & Huang, F. (2024). Qwen2 technical report. *arXiv preprint arXiv:2407.10671*.

Zhou, D.-W., Sun, H.-L., Ning, J., Ye, H.-J., & Zhan, D.-c. (2024). Continual Learning with Pre-Trained Models: A Survey. *ArXiv*, *abs/2401.16386*.

## A  Dataset Samples

Figures 1 shows an example of PsychoLexQA dataset based on psychology document. Figure 2 illustrates an example of PsychoLexQA dataset which has been created using a self-instruct method. Figure 3 shows an example of the PsychoLexEval dataset that includes a multiple-choice question and answer. Figure 4 depicts the process involved in constructing the PsychoLexLLaMA.

---

**Question:** How do researchers measure brain activity during sleep and meditation?

**Answer:** Researchers measure brain activity during sleep and meditation using neuroimaging techniques such as electroencephalograms (EEGs), magnetic resonance imagery (MRI), and positron emission tomography (PET). These techniques record electrical changes in the brain, brain wave patterns, and activation of specific brain areas, allowing researchers to observe and analyze different states of consciousness.

---

Figure 1: Example of PsychoLexQA dataset (Document-based).

---

**Instruct:** Design an experiment to investigate the effects of different types of feedback (positive and negative) on learning outcomes. Include methods for measuring student performance.

**Output:** The experiment will involve three groups of students: one receiving only positive feedback, another receiving only negative feedback, and a third receiving no feedback at all. Each group will engage in a learning task, such as a math problem-solving activity, with their performance assessed through test scores and self-reported motivation levels. The hypothesis posits that students receiving positive feedback will outperform their peers in both task performance and motivation. This experiment will offer insights into the impact of feedback on learning and educational practices.

---

Figure 2: Example of PsychoLexQA dataset (Self-instruct).

---

Which strategy is NOT considered a form of problem-focused coping?

1) Defining the problem
**2) Seeking emotional support**
3) Generating alternative solutions
4) Changing personal goals

Correct Answer is 2.

---

Figure 3: Example of PsychoLexEval dataset.

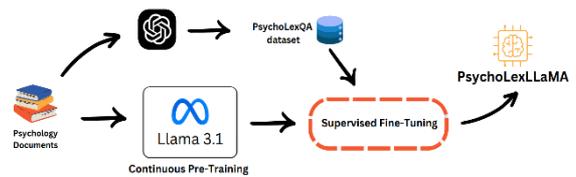

Figure 4: Process of constructing PsychoLexLLaMA model.